\documentclass[runningheads,a4paper]{llncs}

\usepackage{amssymb}
\usepackage{graphicx}

\usepackage{url}
\urldef{\mailsa}\path|{alfred.hofmann, ursula.barth, ingrid.haas, frank.holzwarth,|
\urldef{\mailsb}\path|anna.kramer, leonie.kunz, christine.reiss, nicole.sator,|
\urldef{\mailsc}\path|erika.siebert-cole, peter.strasser, lncs}@springer.com|
\newcommand{\keywords}[1]{\par\addvspace\baselineskip
\noindent\keywordname\enspace\ignorespaces#1}

\begin{document}

\mainmatter  % start of an individual contribution

\title{A Study on Unsupervised Dictionary Learning and Feature Encoding for Action Classification}

\titlerunning{A Study on UDL and Feature Encoding for Action Classification}

\author{Xiaojiang Peng\inst{1,2}%
\and Qiang Peng\inst{1} \and Yu Qiao\inst{2}\and Junzhou Chen\inst{1}\and
Mehtab Afzal\inst{1}}

\institute{$^1$Southwest Jiaotong University, Chengdu, P.R. China\\
$^2$Shenzhen Key Lab of CVPR, Shenzhen Institutes of
Advanced Technology,\\ Chinese Academy of Sciences
}


%\toctitle{Lecture Notes in Computer Science}
%\tocauthor{Authors' Instructions}
\maketitle

\begin{abstract}
Many efforts have been devoted to develop alternative methods to traditional vector quantization in image domain such as sparse coding and soft-assignment. These approaches can be split into a dictionary learning phase and a feature encoding phase which are often closely connected. In this paper, we investigate the effects of these phases by separating them for video-based action classification. We compare several dictionary learning methods and feature encoding schemes through extensive experiments on KTH and HMDB51 datasets. Experimental results indicate that sparse coding performs consistently better than the other encoding methods in large complex dataset (i.e., HMDB51), and it is robust to different dictionaries. For small simple dataset (i.e., KTH) with less variation, however, all the encoding strategies perform competitively. In addition, we note that the strength of sophisticated encoding approaches comes not from their corresponding dictionaries but the encoding mechanisms, and we can just use randomly selected exemplars as dictionaries for video-based action classification.

\end{abstract}

\section{Introduction}
\label{sec:intro}
Much work in computer vision and pattern recognition has applied vector quantization (VQ) to construct mid-level image or video representations. Actually, VQ is a key process of traditional bag-of-words (BoW) model~\cite{sivic2003video}. The framework of BoW mainly contains five steps: feature extraction, dictionary learning, feature encoding, pooling and normalization. As for classical BoW, we usually extract local features from images or videos, learn a dictionary or codebook in training set by clustering algorithm like $K$-means, encode local features to their nearest words (i.e., VQ) and then create a histogram for each image or video by aggregating the frequency of each word.

More recent efforts have been devoted to the dictionary learning and feature encoding processes. An alternative to ``VQ+$K$-means" is sparse coding and its corresponding dictionary which yields better results on image classification benchmark~\cite{yang2009linear}\cite{lee2007efficient}. Sparse representation can be interpreted by the phenomenon in mammalian striate cortex where the receptive fields of simple cells can produce a sparse distribution of output activity in response to natural images~\cite{olshausen1997sparse}.
A natural question is \emph{whether the better performance results from the learned dictionaries or the encoding strategies}.%In other words, is there a better dictionary scheme for feature encoding?

In this paper, we evaluate the effects of unsupervised learned dictionary and feature encoding for video-based action classification. As for video-based action classification, several assessments have been given for the feature extraction~\cite{wang2009evaluation}, feature encoding, pooling and normalizing methods~\cite{wang2012comparative}. To the best of our knowledge, there is still no reported work on the joint evaluation of dictionary learning and feature encoding methods in the context of human action classification.
%The most relative work with ours is \cite{coates2011importance}, but it is limited to image domain.
%In particular, we apply the widely-used spatial-temporal interest points (STIP) with HOG/HOF descriptors~\cite{laptev2008learning} as local video features.
The selected dictionary learning approaches in this paper are namely Random Weights (RW), Random Exemplars (RE), $K$-means, Orthogonal Matching Pursuit (OMP)~\cite{aharon2005k} and Sparse Coding (SC)~\cite{lee2007efficient}. To investigate the performance with different encoding schemes, we employ several advanced feature encoding methods, namely VQ, Soft-assignment (SA)~\cite{liu2011defense}, OMP, SC, Locality-constrained linear coding (LLC)~\cite{wang2010locality}.

The main contributions of this paper come from our analysis of extensive experiments on two popular human action datasets---KTH~\cite{schuldt2004recognizing} and HMDB51~\cite{kuehne2011hmdb}. We give three valuable observations in the context of video-based action classification:
\begin{itemize}
\item All the dictionary learning methods perform similarly on both datasets except for random weights. We can just use random selected exemplars from training set as dictionaries besides those well-designed ones.
\item Sparse coding performs consistently better than the other encoding methods in large complex dataset (i.e., HMDB51), while all the encoding methods perform competitively on small simple dataset.
\item Sparse coding is robust to different dictionaries even that yielded by random weights.
\end{itemize}
These results suggest that we may be able to test new features by employing simple dictionaries and feature encoding schemes (e.g., RE+VQ) when dealing with relatively simple action datasets.

\section{Related Work}
In the past decade, computer vision community has witnessed the popularity and success of the BoW model in many fields, including image classification~\cite{csurka2004visual}, web video categorization~\cite{sivic2003video}, human action recognition~\cite{schuldt2004recognizing}, etc. In classical BoW, vector quantization has been used to assign local features and learn dictionary (i.e., $K$-means, an iterative process of VQ). Actually, VQ maps an input $x$ to a $K$-dimension coded vector $\mathbf{s}$ where the element $s_i$ is 1 if the input belongs to cluster $i$ (i.e., the $i$th word). Both the learning and coding processes are very fast in VQ, but it results in crude representation. Thus, lots of efforts are devoted to improve these two steps. Aharon et al.~\cite{aharon2005k} presented a singular value decomposition ($K$-SVD) based approach to learn effective over-completed dictionary which is a generalized version of $K$-means. Sometimes it is also known as OMP-$K$ since the \emph{Orthogonal Matching Pursuit} is utilized to assign features iteratively at the learning process. Lee et al.~\cite{lee2007efficient} developed a $l_1$ norm based \emph{sparse coding} algorithm, where feature-sign search algorithm was applied for encoding and Lagrange dual method for dictionary learning. Yang et al.~\cite{yang2009linear} employed this SC scheme for image classification and achieved excellent performance. Wang et al.~\cite{wang2010locality} proposed a \emph{locality-constrained linear coding} where a locality constraint was added to the loss function of SC. Zheng et al.~\cite{zheng2011graph} developed a graph regularized sparse coding for image representation which may be hardly practical in video domain due to its expensive time cost. Fisher Kernel coding and GMM were also utilized in image domain~\cite{perronnin2010improving}. \emph{Soft-assignment} is another alterative encoding strategy to VQ which has no corresponding dictionary~\cite{liu2011defense}.

All the dictionary learning schemes mentioned above are unsupervised ones whose loss functions mainly contain reconstruction error and certain unsupervised items. There also exist some supervised dictionary learning methods which are specially designed for classification. Mairal et al.~\cite{mairal2008supervised} proposed to combine the logistic regression with conventional dictionary learning framework. Yang et al.~\cite{yang2010supervised} developed a supervised translation-invariant sparse coding whose purpose is to make the final histogram discriminative under the BoW framework. Jiang et al.~\cite{jiang2011learning} presented a label consistent K-SVD (LC-KSVD) method to learn a discriminative dictionary for sparse coding. Yang et al.~\cite{yang2011fisher} proposed Fisher discrimination dictionary learning (FisherDL) method based on the Fisher criterion to learn a structured dictionary. Supervised dictionary learning strategies usually include discriminative terms which are jointly learned with certain classifiers. Thus, they are much more complex than unsupervised ones.

In our work, we focus on the unsupervised dictionary learning and feature encoding schemes in the BoW model. We note that recent results also implied that the choice of dictionaries may not be as important as one might imagine~\cite{wang2010locality}\cite{coates2011importance}. We extend and complement these findings in video domain with simple and complex datasets. Specially, we provide results using dictionaries generated by random weights, random exemplars, $K$-means, SC and OMP with encoding methods including VQ, SC, OMP, LLC and SA. We will give the formulations of them in the next section.
\section{Method Review}
\begin{figure}[tb]
\centering
\includegraphics[width=0.9\linewidth]{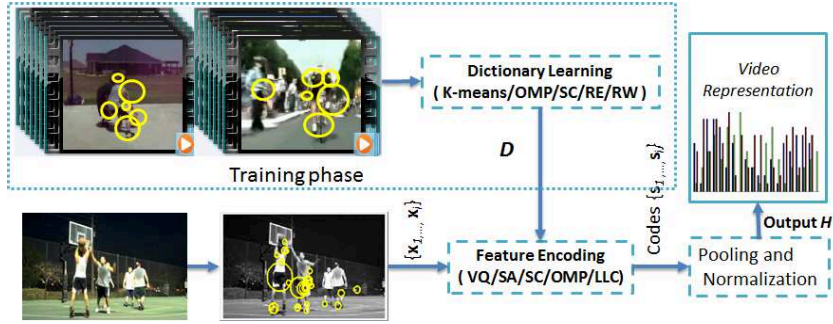}
 \caption{Video representation with the BoW model.}
\label{Fig:framework}
\end{figure}

The typical video representation with the BoW model is shown in Fig.~\ref{Fig:framework}. First, local spatial-temporal features are extracted (e.g., STIP~\cite{laptev2008learning} and Cuboids~\cite{dollar2005behavior}), and then a dictionary is yielded from the features in training set. Next, features are encoded using the learned dictionary, and then all the code coefficients in a single video are pooled and normalized as a histogram, which is used as the final video representation. Here, we focus on the dictionary learning and feature encoding steps. Let $\mathbf{X}=[\mathbf{x}_1, \mathbf{x}_2, ..., \mathbf{x}_N]\in \mathcal{R}^{d\times{N}}$ be a set of feature descriptors, $\mathbf{D}\in \mathcal{R}^{d\times{K}}$ be the learned dictionary and $\mathbf{S}=[\mathbf{s}_1, \mathbf{s}_2, ..., \mathbf{s}_N]\in \mathcal{R}^{K\times{N}}$ be the coefficient vectors. Following are the details of different dictionary learning and feature encoding methods we used.

\subsection{Dictionary Learning Methods}
An unsupervised dictionary learning method aims to yield a certain dictionary $\mathbf{D}=[\mathbf{d}_1, \mathbf{d}_2, ..., \mathbf{d}_K]\in \mathcal{R}^{d\times{K}}$, which can maximally depict the structure of the input feature space. Here, we give the formulations of those methods we used.

\textbf{$K$-means}. It is perhaps the most popular unsupervised way to learn a dictionary due to its simplicity. It aims to minimize the following objective function:
\begin{equation}
C = \sum_{i=1}^K\sum_{j=1}^{n_i}||\mathbf{x}_j^{(i)}-\mathbf{d}_i||_2^2,
\end{equation}
where $\mathbf{x}_j^{(i)}$ denotes the data point included in the cluster $i$ whose center is $\mathbf{d}_i$. In practice, we first select $K$ data points as initial centroids, then assign each data point to the closest centroid and recalculate the positions of the $K$ centroids, repeat these until the centroids no longer move.

\textbf{$K$-SVD} (\textbf{OMP-$K$}). As a generalized version of K-means, \textbf{$K$-SVD}~\cite{aharon2005k} has an alternating objective function:
\begin{eqnarray}
           \min_{\mathbf{D},\mathbf{s}_i}\sum_{i=1}^N||\mathbf{x}_i-\mathbf{D} \mathbf{s}_i||_2^2, \qquad \textrm{s.t.} \quad \forall{i},||\mathbf{x}_i||_{\ell_0}\le K
           \textrm{and} \quad \forall{j},||\mathbf{d}_j||_2^2 =1,
\end{eqnarray}
where $K$ is the largest number of non-zero components in each code $\mathbf{s}_i$. Usually, the codes are computed using OMP algorithm, thus we refer it to as OMP-K here in order to keep pace with its corresponding encoding method. For each single input $\mathbf{x}_i$, OMP greedily selects the most relevant $\mathbf{d}_i$ at each iteration and makes an element of $\mathbf{s}_i$ to be non-zero. After the $K$-th selection, $\mathbf{s}_i$ is updated to minimize $||\mathbf{x}_i-\mathbf{D} \mathbf{s}_i||_2^2$ with allowing only the selected elements to be non-zero. And after computing all the codes, we update the elements of $\mathbf{D}$ one by one via applying SVD to the residual~\cite{aharon2005k}.

\textbf{Sparse Coding}. The standard objective function of $\ell_1$-norm constrained sparse coding for dictionary learning is as follows,
\begin{eqnarray}\label{DL_L1}
 \min_{\mathbf{D},\mathbf{s}_i}\sum_{i=1}^N||\mathbf{x}_i-\mathbf{D} \mathbf{s}_i||_2^2+\lambda \sum_{i}||\mathbf{s}_i||_{\ell_1},
 \qquad \textrm{s.t.} \quad \forall{j},||\mathbf{d}_j||_2^2 =1,
\end{eqnarray}
where $\lambda$ is a sparse factor. Lee et al.~\cite{lee2007efficient} proposed a feature-sign search algorithm to solve the $\ell_1$-regularized least squares (i.e., encoding) problem and a Lagrange dual technique to work out the $\ell_2$-constrained least squares (i.e., dictionary updating). The dictionary is generated iteratively. The main idea of feature-sign search algorithm is to guess the signs of coding coefficients from "helpful" features and then solve an unconstrained quadratic optimization problem.

\textbf{Randomly selected Exemplars} (RE). This approach just fills the columns of $\mathbf{D}$ with normalized vectors randomly sampled from training data set $\mathbf{X}$. This method is absolutely the fastest among all the methods.

\textbf{Random Weights} (RW). Following \cite{coates2011importance}, we also obtain a dictionary by filling the columns of $\mathbf{D}$ with vectors sampled from a unit normal distribution (subsequently normalized to unit length).
\subsection{Feature Encoding Methods}
The purpose of feature encoding is to compute a vector $\mathbf{s}\in \mathcal{R}^{K}$ for input $\mathbf{x}_i$ with $\mathbf{D}$. Here, we give the formulations of the encoding methods we used.

\textbf{Vector quantization} (VQ). VQ is the standard encoding method of BoW, which solves the following constrained objective function:
\begin{equation} %\label{eq:eps},From (\ref{eq:eps}),
\mathbf{s} = arg \min_{\mathbf{s}}||\mathbf{x - Ds}||_2^2, \qquad s.t. ||\mathbf{s}||_{\ell_{0}} = 1,
\end{equation}
where the constraint $||\mathbf{s}||_{\ell_{0}} = 1$ means that there will be only one non-zero element in $\mathbf{s}$, which is found by searching the nearest word in the dictionary.

\textbf{Soft-assignment} (SA). SA means that more than one word will be used. In fact, there are several techniques to realize soft-assignment (e.g., \cite{liu2011defense}, \cite{van2010visual}, \cite{jiang2010representations}). We select the $k$-nearest neighborhood or "localized" version of Liu's~\cite{liu2011defense} (here we name it as SA-K) in our experiments. Let $s_i$ be an element of vector $\mathbf{s}$,
\begin{equation} %\label{eq:eps},From (\ref{eq:eps}),
s_i = \left\{ \begin{array}{ll}
\frac{\exp(-\beta||\mathbf{x}-\mathbf{d}_i||_{2}^{2})}{\sum_{i=1}^K \exp(-\beta||\mathbf{x} - \mathbf{d}_i||_{2}^{2})}; & \textrm{if $\mathbf{d}_i\in N_{k}(\mathbf{x})$},\\
0 & \textrm{otherwise}.
\end{array} \right.
\end{equation}
Where $N_{k}(\mathbf{x})$ denotes the $k$-nearest neighborhood of $\mathbf{x}$. $\beta$ is a smoothing factor controlling the softness of the assignment.

\textbf{Sparse coding}. Given a dictionary \textbf{D}, SC tries to get the code $\mathbf{s}$ for input $\mathbf{x}$ by solving the following function:
\begin{equation}\label{SC}
 \mathbf{s} = arg\min_{\mathbf{s}}||\mathbf{x-Ds}||^2+\lambda ||\mathbf{s}||_{\ell_1},
\end{equation}
This problem is well known as the \emph{lasso} problem~\cite{tibshirani1996regression}. Several algorithms can be used to solve this problem such as least angle regression~\cite{efron2004least} and feature-sign method~\cite{lee2007efficient}. We employ feature-sign scheme here.

\textbf{Orthogonal Matching Pursuit} (OMP-$K$). As mentioned above, given $\mathbf{x}$ and $\mathbf{D}$, we greedily select the most relevant $\mathbf{d}_i$ at each iteration and make an element of $\mathbf{s}$ to be non-zero. After the K-th selection, $\mathbf{s}$ is updated to minimize $||\mathbf{x}-\mathbf{D} \mathbf{s}||_2^2$ by allowing only the selected elements to be non-zero.

\textbf{Locality-constrained linear coding} (LLC). Wang~\cite{wang2010locality} suggested that locality is more essential than sparsity, since
locality must lead to sparsity but not necessary vice versa. The coefficient vector of LLC is obtained by solving the following optimization:
\begin{eqnarray}
\mathbf{s} = arg\min_{\mathbf{s}}||\mathbf{x-Ds}||^2+\lambda||\mathbf{e}\odot\mathbf{s}||^2,
  \qquad \textrm{s.t.} \quad \mathbf{1^{\top}s} = 1,
\end{eqnarray}
where $\mathbf{e}=\exp($dist$(\mathbf{x,D})/\sigma)$ and dist$(\mathbf{x,D})$ denotes the Euclidean distance between $\mathbf{x}$ and $\mathbf{D}$. $\sigma$ is a parameter controlling the weight vector $\mathbf{e}$. In our experiments, we apply the $k$-NN version of LLC (here we call it LLC-$K$), which is an approximation with low computational cost in practice.

\section{Experimental Evaluation and Discussion}
\subsection{Experimental Setting and Datasets}
\begin{figure*}[tb]
\centering
\includegraphics[width=0.95\linewidth]{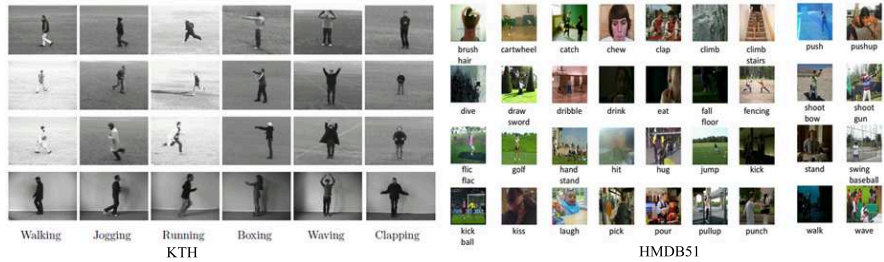}
 \caption{Sample frames from KTH and HMDB51 datasets.}
\label{Fig:Sample}
\end{figure*}

We conduct experiments on two widely-used action datasets, namely ~\cite{schuldt2004recognizing} and HMDB51~\cite{kuehne2011hmdb} datasets. These datasets are collected from controlled experimental setting and web videos.
Some sample frames are illustrated in Fig. \ref{Fig:Sample}.

 The \textbf{KTH} dataset~\cite{schuldt2004recognizing} is one of the most popular datasets in action recognition, which consists of 2391 video clips acted by 25 subjects. It contains 6 action classes: \emph{walking}, \emph{jogging}, \emph{running}, \emph{boxing}, \emph{hand-waving}, and \emph{hand-clapping}. Actions are recorded at 4 environment settings: outdoors, outdoors with camera motion, outdoors with clothing change, and indoors. We follow the experimental settings in~\cite{schuldt2004recognizing} where clips are divided into a training set (16 subjects) and a testing set (9 subjects).

The \textbf{HMDB51} dataset is a large action video database with 51 action categories. Totally, there are 6,766 manually annotated clips which are extracted from a variety of sources ranging from digitized movies to YouTube~\cite{kuehne2011hmdb}. It contains facial actions, general body movements and human interactions. It is a very challenging benchmark due to its high intra-class variation and other fundamental difficulties. In~\cite{kuehne2011hmdb}, three training/testing splits are available, among which we just use the first split to carry out our evaluation.

Spatial-temporal interest points are extracted for both datasets using the version 1.1 of source code from the author's website~\cite{laptev2008learning}. We separate HOG and HOF descriptors, and construct two BoW models. In particular, we randomly sample 100k features to learn dictionaries with size of 4k, respectively. We choose $K=[2, 5, 10]$ for OMP-$K$, SA-$K$ and LLC-$K$, and set the mentioned parameters $[\beta,\lambda,\sigma]$ to be [1, 0.15, 1]. As for dictionary learning phase, we use the VL\_Feat toolbox for K-means, employ the source code from Yang's website~\cite{yang2009linear} for sparse coding, and set 50 iterations for both OMP and SC.  After encoding all the features, we employ sum pooling for all the methods and normalized by ``Power+L2" scheme which demonstrated better results than other post-processing strategies by the previous work of the 3rd author~\cite{wang2012comparative}. Mathematically, sum pooling and ``Power+L2" are as follows,
\begin{itemize}
\item[-]Sum Pooling:  $h_i = \sum_{n=1}^N |s_{ni}|$,
\item[-]Power+L2: $h_i = sign(h_i)|h_i|^\alpha, h_i = h_i/\sqrt{\sum_{i=1}^K|h_i|}$.
\end{itemize}
Where $s_{ni}$ denotes the $i$-th element of the $n$-th code vetor $\mathbf{s}_n$ in a single video.
We set $\alpha$ to be 0.5 in our experiments. For classification, we first compute RBF kernels with $\chi^2$ distance for both HOG and HOF channels~\cite{laptev2008learning}, and then get the average kernel as inputs for kernel SVM classifier. As for multi-class classification, we use the \emph{one-against-rest} approach and select the class with the highest score.

\subsection{Evaluation on KTH}
\begin{figure*}[t]
\centering
\includegraphics[width=0.8\linewidth]{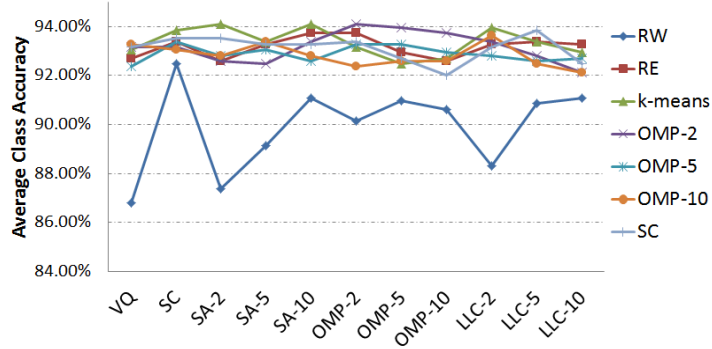}
 \caption{The corresponding graph of Table \ref{tab:resultsKTH}.}
\label{Fig:resultsKTH}
\end{figure*}

Our first purpose is to explore the effects of different dictionary learning and feature encoding methods on the relatively simple dataset. We test all the probable combinations among different dictionaries and encoding schemes.
Table \ref{tab:resultsKTH} shows the results on the KTH dataset. To make the results distinctive, we illustrate them in Fig. \ref{Fig:resultsKTH}.

From the results in Table \ref{tab:resultsKTH}, we confirm that it is no need to match one dictionary with specific encoding scheme.
From Fig. \ref{Fig:resultsKTH}, we note that all the dictionaries except \emph{Random Weights} with various feature encoding methods achieve similar results since the differences among them are no more than 2\%.
We interpret that dictionary using random weights has little choice to tile the space of the input features, thus the KNN-like encoding methods (i.e., those we used except for SC) make no sense when choosing their neighbors. However, sparse coding achieves competitive result even using random weights as its dictionary. It may be explained that SC is optimized by the ``helpful" orientations of all the coordinates of $\mathbf{s}$. The top 3 results are set in bold as shown in Table \ref{tab:resultsKTH}.

\begin{table}[t]
\caption{Results of different dictionaries and encoding schemes on the KTH dataset. All the numbers are percent accuracy.}
\label{tab:resultsKTH}
\centering
\begin{tabular}{c||ccccccc}\hline
Methods & RW & RE & $K$-means & OMP-2 & OMP-5 & OMP-10 & SC \\\hline\hline
VQ      &86.80&	92.70&	93.05&	93.17&	92.36&	93.29&	93.17\\
OMP-2      &90.16&	93.75&	93.17&	\textbf{94.10}&	93.29&	92.36&	93.40\\
OMP-5      &90.97&	92.94&	92.48&	\textbf{93.98}&	93.29&	92.59&	92.71\\
OMP-10      &90.62&	92.59&	92.71&	93.75&	92.94&	92.59&	92.01\\
SA-2      &87.38&	92.59&	\textbf{94.09}&	92.59&	92.82&	92.82&	93.52\\
SA-5      &89.12&	93.28&	93.40&	92.48&	93.06&	93.40&	93.29 \\
SA-10      &91.09&	93.75&	\textbf{94.10}&	93.40&	92.59&	92.82&	93.29 \\
LLC-2      &88.31&	93.29&	\textbf{93.98}&	93.40&	92.82&	93.63&	93.17 \\
LLC-5      &90.86&	93.40&	93.40&	92.82&	92.59&	92.48&	93.87 \\
LLC-10      &91.09&	93.29&	92.94&	92.13&	92.71&	92.13&	92.48 \\
SC      &\emph{92.48}&	93.40&	93.87&	93.17&	93.40&	93.06&	93.52 \\\hline
\end{tabular}
\end{table}

\subsection{Evaluation on HMDB51}
\begin{figure}[t]
\centering
\includegraphics[width=0.8\linewidth]{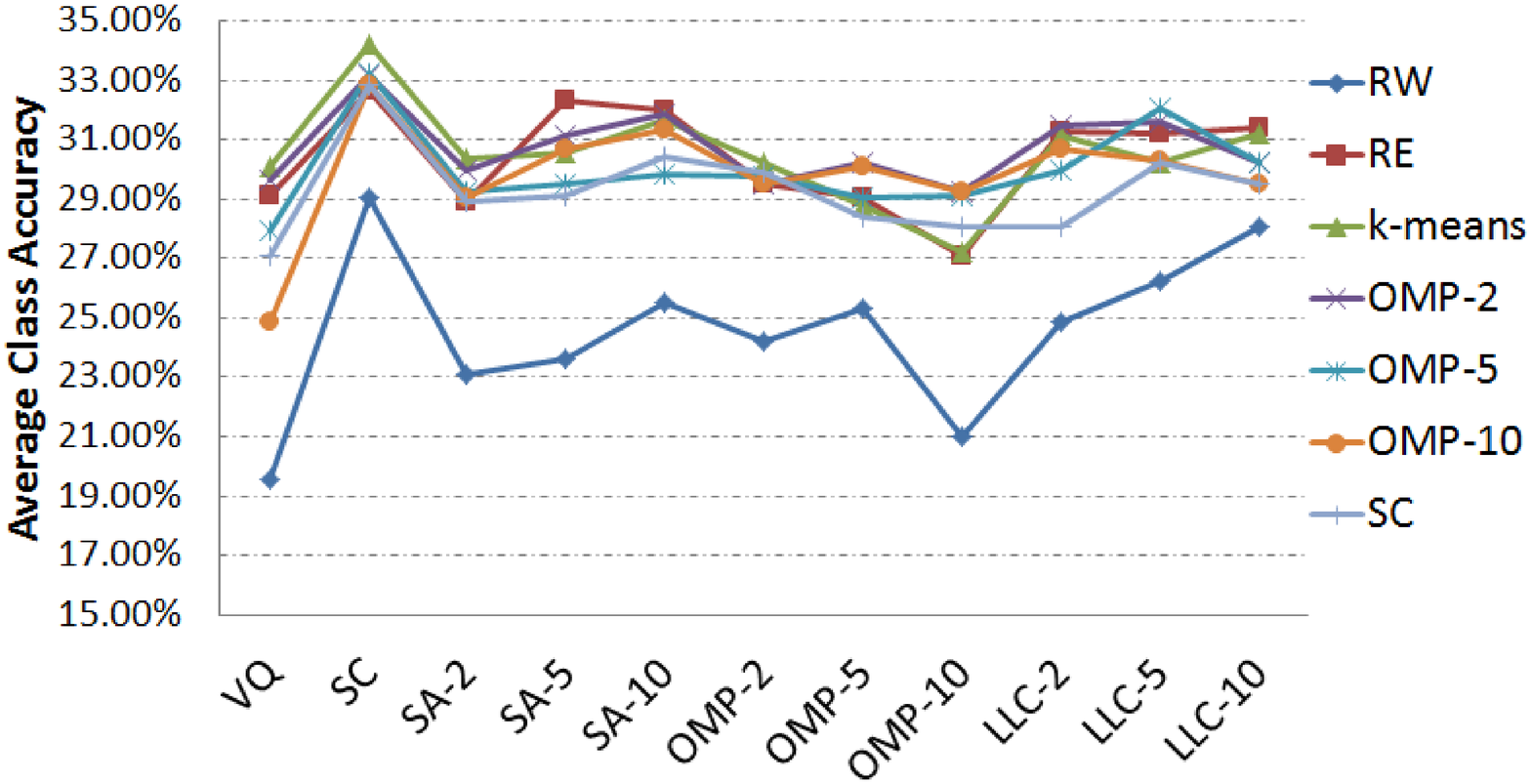}
 \caption{The corresponding graph of Table \ref{tab:resultsHMDB51}.}
\label{Fig:resultsHMDB51}
\end{figure}

We perform similar experiments on the HMDB51 dataset to check and extend our previous observations. The results are exhibited in Table \ref{tab:resultsHMDB51} and Fig. \ref{Fig:resultsHMDB51}.

As shown in Table \ref{tab:resultsHMDB51} and Fig. \ref{Fig:resultsHMDB51}, feature encoding using SC obtains consistent better results than the others we used, with 3.52\%-9.54\% more than traditional VQ.  And if we look at the row of SC, it is clear that the power of SC on HMDB51 comes not from the dictionaries, but from the SC mechanism itself since the best result is not occurred in the ``SC+SC" combination. The top 3 results on the HMDB51 are set in bold as shown in Table \ref{tab:resultsHMDB51}.

\begin{table}[t]
\caption{Results of the 1st training/testing split on the HMDB51 dataset. All the numbers are percent accuracy.}
\label{tab:resultsHMDB51}
\centering
\begin{tabular}{c||ccccccc}\hline
Methods & RW & RE & $K$-means & OMP-2 & OMP-5 & OMP-10 & SC \\\hline\hline
VQ      &19.54&	29.15&	30.13&	29.67&	27.91&	24.84&	27.12\\
OMP-2      &24.18&	29.48&	30.26&	29.54&	29.80&	29.48&	29.93\\
OMP-5      &25.29&	29.08&	28.76&	30.26&	29.08&	30.07&	28.43\\
OMP-10      &20.98&	27.06&	27.25&	29.28&	29.15&	29.22&	28.04\\
SA-2      &23.07&	28.89&	30.39&	30.00&	29.22&	29.08&	28.95\\
SA-5      &23.59&	32.35&	30.59&	31.18&	29.48&	30.72&	29.15 \\
SA-10      &25.49&	32.03&	31.70&	31.90&	29.87&	31.37&	30.46 \\
LLC-2      &24.84&	31.31&	31.18&	31.50&	30.00&	30.72&	28.10\\
LLC-5     &26.21&	31.24&	30.26&	31.63&	32.09&	30.33&	30.20\\
LLC-10    &28.04&	31.44&	31.24&	30.26&	30.26&	29.54&	29.48 \\
SC     &\emph{29.08}&	32.67&	\textbf{34.25}&	\textbf{33.20}&	\textbf{33.27}&	32.88&	32.88 \\\hline
\end{tabular}
\end{table}
\subsection{Discussion}
\textbf{Methods and datasets}.
From all the results on both datasets, we find several observations. First, comparing all the column-wise results in both Table \ref{tab:resultsKTH} and \ref{tab:resultsHMDB51}, the efficient and economical dictionary is perhaps the one yielded by $K$-means algorithm since the best results commonly exist in those columns. Second, considering accuracies of different feature encoding schemes, sparse coding is significantly better than the others on large dataset and it is robust to all the dictionaries even random weights are used. Third, there is no evident improvement or deterioration by increasing the number of neighbors for OMP, SA, and LLC in both encoding and learning phases. Finally, it is obvious that the strength of well-designed encoding strategies is hardly related to the learned dictionaries but the encoding mechanisms.

The distinctions between the KTH dataset and HMDB51 dataset come from the complexity of action videos and the number of categories. KTH owns only 6 classes and the STIPs are well detected at the motion foreground. It is large enough to tile the feature space by a dictionary with the size of 4,000. That means about 667 words can be assigned to each class averagely. Thus, no matter the encoding scheme is, the distributions of all the histograms for each class can be separated effectively.
%The statistical frequency of each words (HOF dictionary) for each action class generated by several encoding methods with K-means are shown in Fig.....
In these cases, all the approaches perform competitively.
The inverse case is for the HMDB51 dataset, the STIPs are not only extracted in motion foreground but also in the background due to serious camera motion and it is not enough to cover the feature space for each class with less than 80 words because of large variations. Thus, serious overlaps among the word distributions of different classes occur on HMDB51 dataset, while sparse coding can slightly alleviate these overlaps.
\begin{table}[t]
\caption{Results of different dictionary sizes on the KTH dataset. All the numbers are percent accuracy.}
\label{tab:Dsize}
\centering
\begin{tabular}{rc||ccccc}\hline
Encoder/Dictionary& \rotatebox{90}{Size} & 100 & 200 & 500 & 1000 & 4000\\\hline\hline
VQ/RE & & 90.74 & 90.28 & 92.82 & 92.94 & 92.70\\
VQ/$K$-means & & 90.05 & 91.09 & 91.90 & 93.52 & 93.05\\
SC/RE & & 91.32 & 91.20 & 93.17 & 92.71 & 93.40\\
SC/$K$-means & & 90.74 & 92.36 & 93.06 & 93.05 & 93.87\\
SC/SC & & 90.74 & 91.55 & 92.24 & 92.71 & 93.52\\
SA-5/$K$-means & & 90.05 & 91.32 & 90.97 & 93.17 & 93.40\\
OMP-5/OMP-5 & & 89.23 & 89.00 & 90.97 & 92.31 & 93.29\\\hline
\end{tabular}
\end{table}

\begin{table}[t]
\caption{The cost of different dictionary learning methods with the size of 4k on KTH dataset.}
\label{tab:costdictionary}
\centering
\begin{tabular}{cccccccc}\hline
Methods & RW & RE & $K$-means & OMP-2 & OMP-5 & OMP-10 & SC \\\hline
HOG     &0.15s&	0.15s&	7.48min &	1.45h&	2.83h &	5.50h&	23.14h\\
HOF      &0.18s&	0.18s&	7.50min &	1.75h&	3.05h&	5.94h&	31.47h\\\hline
\end{tabular}
\end{table}

\begin{table}[t]
\caption{The cost of different feature encoding methods for a single input $\mathbf{s}$ on KTH, where the dictionary is fixed by $K$-means with the size of 4k. All the numbers are in microsecond.}
\label{tab:costEncoding}
\centering
\begin{tabular}{cccccccccccc}\hline
 Methods& VQ & \rotatebox{90}{SA-2} & \rotatebox{90}{SA-5} & \rotatebox{90}{SA-10} & \rotatebox{90}{LLC-2} & \rotatebox{90}{LLC-5} & \rotatebox{90}{LLC-10} & \rotatebox{90}{OMP-2} & \rotatebox{90}{OMP-5} & \rotatebox{90}{OMP-10} & SC \\\hline
HOG     &0.31&	0.48&	0.49 &	0.48&	0.72 & 0.68& 0.68&1.40 &	1.86& 3.31&	7.9\\
HOF      &0.25&	0.45&	0.44 &	0.44& 0.65&0.67 & 0.62 &1.15& 2.05 &	4.08&7.3\\\hline
\end{tabular}
\end{table}
\textbf{Dictionary size}.
We further discuss the properties of dictionary by using different dictionary sizes on the KTH dataset. The results of several selected methods with different dictionary sizes are shown in Table \ref{tab:Dsize}. Generally, the results are improved with the dictionary size. It seems that an appropriate size of dictionary is 4k or larger. When comparing the results per column, we observe that all the methods perform similarly. That means our previous conclusions are independent with dictionary size.

\textbf{Cost}. For the dictionary learning methods, the time consumptions are ranked as: RE$\approx$RW$\ll K$-means$\ll$OMP-2$<$OMP-5$<$OMP-10$<$SC. RE and RW are the fastest ones obviously, and the others have to spend expensive time cost for optimization. As for the encoding approaches, the order is VQ$<$SA$<$LLC$<$OMP$<$SC, and the cost increases with the number K. Table \ref{tab:costdictionary} shows the detailed costs of dictionary learning schemes. We also test the costs of feature encoding methods by 50 randomly selected videos from the KTH dataset, where the dictionaries are yielded by $K$-means. The average costs of encoding strategies for one single input $\mathbf{x}$ (i.e., HOG or HOF) are illustrated in Table \ref{tab:costEncoding}.
The runtime is obtained on an Acer laptop with a 2.5 GHz Intel Core i5 CPU and 4 GB RAM. And we set 50 iterations for OMP and SC. It is worth noting that the cost of encoding by SC is related to dictionary. For instance, when the random weights are used, sparse coding is very time-consuming because active set would become large and the feature-sign searching process may be hard to terminate.

%\begin{table*}[!htb]
%% increase table row spacing, adjust to taste
%\renewcommand{\arraystretch}{1.3}
% %if using array.sty, it might be a good idea to tweak the value of
%% \extrarowheight as needed to properly center the text within the cells
%\caption{An Example of a Table}
%\label{table_example}
%\centering
%% Some packages, such as MDW tools, offer better commands for making tables
%% than the plain LaTeX2e tabular which is used here.
%\begin{tabular}{|c||c|c|c|c|c|c|c|c|c|c|c|c|c|c|c|c|c|c|c|c|}
%\hline
%One & Two& Two& Two& Two& Two& Two& Two& Two& Two& Two& Two& Two& Two& Two& Two& Two& Two& Two& Two& Two\\
%\hline
%Three & Four& Two& Two& Two& Two& Two& Two& Two& Two& Two& Two& Two& Two& Two& Two& Two& Two& Two& Two& Two\\
%\hline
%\end{tabular}
%\end{table*}
\section{Conclusion}
In this paper, we have conducted extensive experiments to compare the performance of several unsupervised dictionary learning and encoding schemes on KTH and HMDB51 datasets. Experimental results demonstrate that sparse coding performs consistently better than the other encoding methods on the HMDB51 dataset, and it is robust to different dictionaries. As for the KTH dataset, all the encoding strategies perform competitively due to its small variations. By decoupling the learning and encoding phases of sparse coding, we discovered that
the strength mainly comes from the encoding mechanism. This is an extended observation in the context of video-based action classification, and it shows the trend for further studies and improvements in the BoW model.


%\subsubsection{Acknowledgment.}
%This work is partly supported by the 2013 Doctoral Innovation Funds of Southwest Jiaotong University, the National Natural Science Foundation of China (No.61003143) and the Fundamental Research Funds for Central Universities (No.SWJTU12CX094).

%\begin{thebibliography}{4}
%
%\bibitem{jour} Smith, T.F., Waterman, M.S.: Identification of Common Molecular
%Subsequences. J. Mol. Biol. 147, 195--197 (1981)
%
%\bibitem{lncschap} May, P., Ehrlich, H.C., Steinke, T.: ZIB Structure Prediction Pipeline:
%Composing a Complex Biological Workflow through Web Services. In: Nagel,
%W.E., Walter, W.V., Lehner, W. (eds.) Euro-Par 2006. LNCS, vol. 4128,
%pp. 1148--1158. Springer, Heidelberg (2006)
%
%\bibitem{book} Foster, I., Kesselman, C.: The Grid: Blueprint for a New Computing
%Infrastructure. Morgan Kaufmann, San Francisco (1999)
%
%\bibitem{proceeding1} Czajkowski, K., Fitzgerald, S., Foster, I., Kesselman, C.: Grid
%Information Services for Distributed Resource Sharing. In: 10th IEEE
%International Symposium on High Performance Distributed Computing, pp.
%181--184. IEEE Press, New York (2001)
%
%\bibitem{proceeding2} Foster, I., Kesselman, C., Nick, J., Tuecke, S.: The Physiology of the
%Grid: an Open Grid Services Architecture for Distributed Systems
%Integration. Technical report, Global Grid Forum (2002)
%
%\bibitem{url} National Center for Biotechnology Information, \url{http://www.ncbi.nlm.nih.gov}
%
%\end{thebibliography}
\bibliographystyle{splncs}
\bibliography{pengxj_arxiv2013_UDL_Encoding}

\begin{thebibliography}{10}

\bibitem{sivic2003video}
Sivic, J., Zisserman, A.:
\newblock Video google: A text retrieval approach to object matching in videos.
\newblock In: Computer Vision, IEEE International Conference on. (2003)
  1470--1477

\bibitem{yang2009linear}
Yang, J., Yu, K., Gong, Y., Huang, T.:
\newblock Linear spatial pyramid matching using sparse coding for image
  classification.
\newblock In: Computer Vision and Pattern Recognition, IEEE Conference on.
  (2009)  1794--1801

\bibitem{lee2007efficient}
Lee, H., Battle, A., Raina, R., Ng, A.Y.:
\newblock Efficient sparse coding algorithms.
\newblock Advances in neural information processing systems \textbf{19} (2007)
  801

\bibitem{olshausen1997sparse}
Olshausen, B., Field, D.,  et~al.:
\newblock Sparse coding with an overcomplete basis set: A strategy employed by
  vi?
\newblock Vision research \textbf{37}(23) (1997)  3311--3326

\bibitem{wang2009evaluation}
Wang, H., Ullah, M.M., Klaser, A., Laptev, I., Schmid, C.,  et~al.:
\newblock Evaluation of local spatio-temporal features for action recognition.
\newblock In: British Machine Vision Conference (BMVC). (2009)

\bibitem{wang2012comparative}
Wang, X., Wang, L., Qiao, Y.:
\newblock A comparative study of encoding, pooling and normalization methods
  for action recognition.
\newblock In: Asian Conference on Computer Vision (ACCV). (2012)

\bibitem{aharon2005k}
Aharon, M., Elad, M., Bruckstein, A.:
\newblock K-svd: an algorithm for designing of overcomplete dictionaries for
  sparse representation.
\newblock IEEE Trans. On Signal Processing \textbf{54}(11) (2005)  4311--5322

\bibitem{liu2011defense}
Liu, L., Wang, L., Liu, X.:
\newblock In defense of soft-assignment coding.
\newblock In: Computer Vision, IEEE International Conference on. (2011)
  2486--2493

\bibitem{wang2010locality}
Wang, J., Yang, J., Yu, K., Lv, F., Huang, T., Gong, Y.:
\newblock Locality-constrained linear coding for image classification.
\newblock In: Computer Vision and Pattern Recognition, IEEE Conference on.
  (2010)  3360--3367

\bibitem{schuldt2004recognizing}
Schuldt, C., Laptev, I., Caputo, B.:
\newblock Recognizing human actions: A local svm approach.
\newblock In: Pattern Recognition, Proceedings of the 17th International
  Conference on. Volume~3. (2004)  32--36

\bibitem{kuehne2011hmdb}
Kuehne, H., Jhuang, H., Garrote, E., Poggio, T., Serre, T.:
\newblock Hmdb: A large video database for human motion recognition.
\newblock In: Computer Vision, IEEE International Conference on. (2011)
  2556--2563

\bibitem{csurka2004visual}
Csurka, G., Dance, C., Fan, L., Willamowski, J., Bray, C.:
\newblock Visual categorization with bags of keypoints.
\newblock In: Workshop on statistical learning in computer vision, ECCV.
  Volume~1. (2004)  22--37

\bibitem{zheng2011graph}
Zheng, M., Bu, J., Chen, C., Wang, C., Zhang, L., Qiu, G., Cai, D.:
\newblock Graph regularized sparse coding for image representation.
\newblock Image Processing, IEEE Trans. on \textbf{20}(5) (2011)  1327--1336

\bibitem{perronnin2010improving}
Perronnin, F., S{\'a}nchez, J., Mensink, T.:
\newblock Improving the fisher kernel for large-scale image classification.
\newblock In: Computer Vision--ECCV. (2010)  143--156

\bibitem{mairal2008supervised}
Mairal, J., Bach, F., Ponce, J., Sapiro, G., Zisserman, A.:
\newblock Supervised dictionary learning.
\newblock Advances in neural information processing systems (2008)

\bibitem{yang2010supervised}
Yang, J., Yu, K., Huang, T.:
\newblock Supervised translation-invariant sparse coding.
\newblock In: Computer Vision and Pattern Recognition, IEEE Conference on.
  (2010)  3517--3524

\bibitem{jiang2011learning}
Jiang, Z., Lin, Z., Davis, L.S.:
\newblock Learning a discriminative dictionary for sparse coding via label
  consistent k-svd.
\newblock In: Computer Vision and Pattern Recognition, IEEE Conference on.
  (2011)  1697--1704

\bibitem{yang2011fisher}
Yang, M., Zhang, L., Feng, X., Zhang, D.:
\newblock Fisher discrimination dictionary learning for sparse representation.
\newblock In: Computer Vision, IEEE International Conference on. (2011)
  543--550

\bibitem{coates2011importance}
Coates, A., Ng, A.Y.:
\newblock The importance of encoding versus training with sparse coding and
  vector quantization.
\newblock In: International conference on machine learning (ICML). Volume~8.
  (2011) ~10

\bibitem{laptev2008learning}
Laptev, I., Marszalek, M., Schmid, C., Rozenfeld, B.:
\newblock Learning realistic human actions from movies.
\newblock In: Computer Vision and Pattern Recognition, IEEE Conference on.
  (2008)  1--8

\bibitem{dollar2005behavior}
Doll{\'a}r, P., Rabaud, V., Cottrell, G., Belongie, S.:
\newblock Behavior recognition via sparse spatio-temporal features.
\newblock In: Visual Surveillance and Performance Evaluation of Tracking and
  Surveillance, 2nd Joint IEEE International Workshop on. (2005)  65--72

\bibitem{van2010visual}
van Gemert, J., Veenman, C., Smeulders, A., Geusebroek, J.:
\newblock Visual word ambiguity.
\newblock PAMI, IEEE Trans. on \textbf{32}(7) (2010)  1271--1283

\bibitem{jiang2010representations}
Jiang, Y., Yang, J., Ngo, C., Hauptmann, A.:
\newblock Representations of keypoint-based semantic concept detection: A
  comprehensive study.
\newblock Multimedia, IEEE Trans. on \textbf{12}(1) (2010)  42--53

\bibitem{tibshirani1996regression}
Tibshirani, R.:
\newblock Regression shrinkage and selection via the lasso.
\newblock Journal of the Royal Statistical Society. Series B (Methodological)
  (1996)  267--288

\bibitem{efron2004least}
Efron, B., Hastie, T., Johnstone, I., Tibshirani, R.:
\newblock Least angle regression.
\newblock The Annals of statistics \textbf{32}(2) (2004)  407--499

\end{thebibliography}

%\section*{Appendix: Springer-Author Discount}
%
%LNCS authors are entitled to a 33.3\% discount off all Springer
%publications. Before placing an order, the author should send an email,
%giving full details of his or her Springer publication,
%to \url{orders-HD-individuals@springer.com} to obtain a so-called token. This token is a
%number, which must be entered when placing an order via the Internet, in
%order to obtain the discount.
%
%\section{Checklist of Items to be Sent to Volume Editors}
%Here is a checklist of everything the volume editor requires from you:
%
%
%\begin{itemize}
%\settowidth{\leftmargin}{{\Large$\square$}}\advance\leftmargin\labelsep
%\itemsep8pt\relax
%\renewcommand\labelitemi{{\lower1.5pt\hbox{\Large$\square$}}}
%
%\item The final \LaTeX{} source files
%\item A final PDF file
%\item A copyright form, signed by one author on behalf of all of the
%authors of the paper.
%\item A readme giving the name and email address of the
%corresponding author.
%\end{itemize}
\end{document}